\documentclass[conference]{IEEEtran}
\IEEEoverridecommandlockouts
\usepackage{cite}
\usepackage{amsmath,amssymb,amsfonts}
\usepackage{algorithmic}
\usepackage{graphicx}
\usepackage{subfig} 

\usepackage{textcomp}
\usepackage{url}
\begin{document}

\title{CORNET 2.0: A Co-Simulation Middleware for Robot Networks\\
}

\author{\IEEEauthorblockN{1\textsuperscript{st} Srikrishna Acharya}
\IEEEauthorblockA{\textit{RBCCPS\thanks{\textsuperscript{1} Robert Bosch Center for Cyber Physical Systems, IISc}\footnotemark\textsuperscript{1}}\\
\textit{Indian Institute of Science}\\ \text
Bengaluru, India \\
bacharya@iisc.ac.in}
\and
\IEEEauthorblockN{2\textsuperscript{nd} Bharadwaj Amrutur}
\IEEEauthorblockA{\textit{RBCCPS\footnotemark \textsuperscript{1},ECE,ARTPARK\footnotemark \textsuperscript{2}}\\
	\textit{Indian Institute of Science}\\
	Bengaluru, India \\
	amrutur@iisc.ac.in}
\and
\IEEEauthorblockN{3\textsuperscript{rd} Mukunda Bharathesa}
\IEEEauthorblockA{\textit{Artpark\footnotemark\textsuperscript{2}/RBCCPS\footnotemark\textsuperscript{1}\thanks{\textsuperscript{2}AI \& Robotics Technology Park, IISc} }\\
\textit{Indian Institute of Science}\\
Bengaluru, India \\
mukunda@artpark.in}
\and
\IEEEauthorblockN{4\textsuperscript{th} Yogesh Simmhan}
\IEEEauthorblockA{\textit{CDS\thanks{\textsuperscript{3} Department of Computational and Data Sciences (CDS), IISc}\footnotemark\textsuperscript{3} and RBCCPS\footnotemark\textsuperscript{1}} \\
\textit{Indian Institute of Science}\\
Bengaluru, India \\
simmhan@iisc.ac.in}
}

\maketitle

\begin{abstract}
We present a networked co-simulation framework for multi-robot systems applications. We require a simulation framework that captures both physical interactions and communications aspects to effectively design such complex systems. This is necessary to co-design the multi-robots' autonomy logic and the communication protocols. The proposed framework extends existing tools to simulate the robot’s autonomy and network-related aspects. We have used Gazebo with ROS/ROS2 to develop the autonomy logic for robots and mininet-WiFi as the network simulator to capture the cyber-physical systems properties of the multi-robot system. This framework addresses the need to seamlessly integrate the two simulation environments by synchronizing mobility and time, allowing for easy migration of the algorithms to real platforms. The framework supports container-based virtualization and extends a generic robotic framework by decoupling the data plane and control plane.   

\end{abstract}


\section{Introduction}
Autonomous robots are game-changers in domains as diverse as transportation, disaster response, mining, exploration and health care. With the technological advances in artificial intelligence, cloud-edge computing, and machine learning, autonomous robots are expected to play a vital role in the future of civic transportation systems. With the emergence of 5G and vehicle-to-infrastructure (V2X) frameworks, use-cases like infrastructure assistance~\cite{tsukada:hal-01558066} and advanced driver assistance systems that are situation aware~\cite{machines5010006} are enabled. Implementation of these frameworks requires development of new algorithms and proper tuning of parameters. However, it is not always feasible to use real robots when iteratively designing and validating these algorithms as it is time-consuming and costly, with both hardware and environment in the loop.  Therefore it is necessary to accurately simulate the system with appropriate tools.

However, most of the R\&D efforts into simulators address two distinct and separate areas. Network simulators like NS3\cite{ns3}, OMNET++\cite{opnet}, etc., attempt to represent the traffic over the communication network, while autonomous systems simulators like Gazebo\cite{koenig2004design}, AirSim\cite{shah2018airsim}, etc., focus on the physical interactions of the robot with their environment. There is a dearth of open-source tools to simulate multi-agent systems with realistic control-loop interactions and detailed network-side simulation. In this work, we address this gap through an integrated framework for the realistic simulation of connected autonomous robots, and extend our earlier work on CORNET\cite{acharya2020cornet}. Physic simulators employ a continuous time-based simulation strategy, whereas network simulators are event-driven. Due to the different design principles behind physics and network simulators, synchronization mechanisms needs meticulous planning. 

The main contributions of this paper are as follows:
\begin{itemize}
    \item Integration design and implementation of two open-source simulation engines, Gazebo-ROS, and mininet-WiFi\cite{fontes2015mininet}, with mobility and time synchronization;
    \item Extension of the framework to support any robotic middleware like ROS, ROS2, etc.; 
    \item Support for container-based virtualization for better isolation of network and application logic.
\end{itemize}

The CORNET 2.0 source code is open-sourced under MIT License and available on GitHub\cite{cornet2.0}. 

\begin{figure}[t]
\centering
	\includegraphics[width=.8\columnwidth]{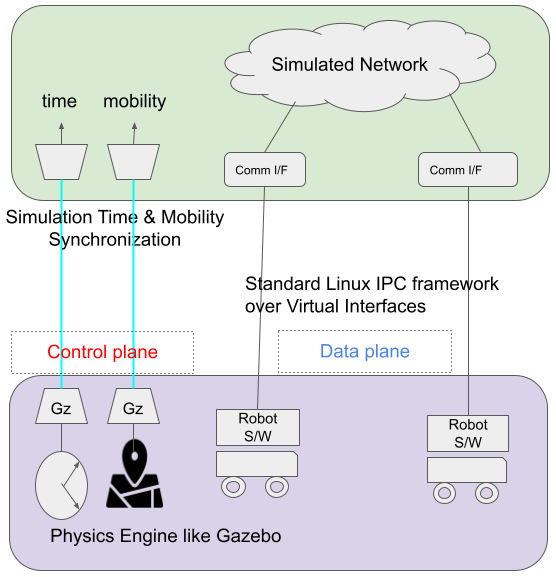}
	\caption{Architecture of CORNET 2.0}
	\label{fig:overviewcornet2}
\end{figure}

\section{Related Work}
In this section, we provide a summary of existing joint robot-network simulators. Due to their popularity, most frameworks in literature like FlyNetSim\cite{baidya2018flynetsim}, CUSCUS\cite{zema2018cuscus}, AVENS\cite{marconato2017avens}, and CORNET 1.0\cite{acharya2020cornet}, focus on UAV-centric applications.
These frameworks are designed for a specific application and do not translate appropriately to other robotic applications. For example, FlyNetSim and CORNET use dronekit-python and MAVROS, respectively, for control messages and are application-specific implementations. Moreover, in CUSCUS and AVENS, only the UAV positions are only accounted for and they do not capture the target application’s network traffic.
 
On the other hand, some frameworks like RoboNetSim\cite{kudelski2013robonetsim} and RosNetSim\cite{calvo2021ros} are designed with general-purpose robotics in mind. RoboNetSim integrates ARGoS with NS2 and provides a synchronization mechanism bridging the physics and network simulators’ continuous-time and discrete-event natures.  However, it lacks a proper packet capture mechanism for target applications. Another notable framework RosNetSim addresses many of the above mentioned drawbacks. ROSNETSIM is built on ROS framework and provides a tunable simulation description agnostic to both network and physics simulator. It extracts geometrical information from the physics simulator to be utilized by the network simulator, which is a useful capability. Unavoidably, the framework is tightly coupled with the ROS framework, and network side implementations must be explicitly implemented.  A brief comparison is provided in Table \ref{tab:table1}. 
  
\begin{table}[t]
	\begin{center}
		\caption{Comparison of Joint Simulators}
		\label{tab:table1}
		\begin{tabular}{p{0.06\textwidth} p{0.03\textwidth} p{0.06\textwidth} p{0.06\textwidth} p{0.16\textwidth}}
			\textbf{Simulator}& \textbf{Open-Source?}& \textbf{Network}& \textbf{Physics}& \textbf{Comments} \\ \hline
			CUSCUS & No & NS-3  & FL-AIR & Specific for drones, no data path, and no synchronization policy   \\ \hline
		    AVENS & Yes & OMNET++ & X-Plane  & Specific for drones, XML based update and no data path \\ \hline
			FlyNetSim & Yes & NS3 & DroneKit & Specific for drones \\ \hline
			RosNetSim & Yes & -  & - & Tied to ROS framework, synchronization limited to window size. \\ \hline
			RoboNetSim & Yes & NS2/NS3 & ARGoS  & No Datapath \\ 
			CORNET & Yes & NS3 & Gazebo  & Specific to drones with tightly coupled with MAVROS  \\ 
            CORNET 2.0 & Yes & mininet-WiFi & Gazebo  & Generic framework decoupling data plane and control plane  \\ 
		\end{tabular}
	\end{center}
\end{table}

\section{CORNET 2.0 Co-simulation Architecture}
\begin{figure}[t]
\centering
	\includegraphics[width=1.0\columnwidth]{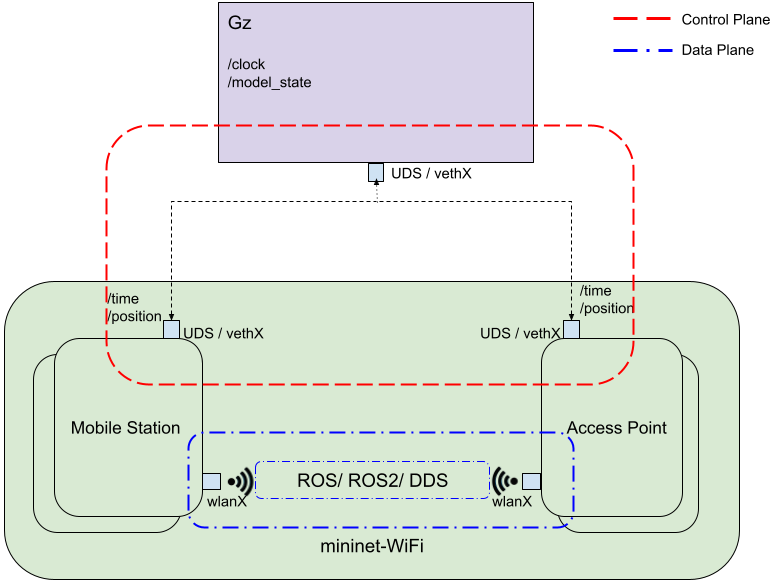}
	\caption{CORNET 2.0 Interaction Pathways}
	\label{fig:cornet2}
\end{figure}

Fig. \ref{fig:overviewcornet2} provides an overview of the entities within CORNET 2.0 framework while Fig. \ref{fig:cornet2}  illustrates the interaction details. We use Gazebo to realize the robot simulation’s physics aspects and mininet-WiFi with Containernet\cite{peuster2016medicine} support to realize the networking aspects. There is a data plane to interconnect the robot's software processes and a control plane to pass the simulation time and the robot's positions from the physics simulator into the network simulator. The data plane interconnection is enabled through virtual interfaces, which allows for any inter-robot communication protocol to be used by the application under simulation. The CORNET 2.0 middleware allows seamless integration of the two simulation tools, providing an inter-simulator datapath, mobility, and time synchronization. In the following subsections, we elaborate on each component of the framework.

\subsection{Physics Simulator}
Gazebo is an open-source robotic simulator, which provides a physics engine, high-quality graphics rendering, UI, and other sensor models. 
It uses an Simulation Description Format (SDF) format that defines the environment and inertial properties of the robots. Robot Operating System (ROS/ROS2) is a set of open-source software libraries and tools that has been created to help developers to build robotic applications. \texttt{gazebo\_ros\_pkgs} is a set of packages used for integrating the Gazebo simulation with either version of ROS. They provide the bridge between Gazebo’s API and Transport systems and ROS messages and services. 

Many modern platforms like UE\cite{sanders2016introduction}, AirSIM, and Unity\cite{haas2014history} offer an astounding level of realistic representation of physics and dynamics of the robot and its interactions with the environment. But our choice of Gazebo as the default physical system simulator for CORNET 2.0 is primarily based on its native support in Linux environments and the extensive community support. That said, the CORNET 2.0 middleware is not tightly coupled with  Gazebo, and any physical system simulator which provides the clock source and states of the robot can easily be integrated with our framework.

\subsection{Network Simulator}
Mininet-WiFi extends the mininet network simulator to support WiFi using a \texttt{mac80211\_hwsim} simulator to simulate an arbitrary number of IEEE 802.11 radios for mac80211. It uses the \texttt{wmediumd} library to emulate wireless medium. Specifically:
\begin{itemize}
	\item It implements a back-off algorithm to emulate the CSMA/CA protocol
	\item It allows an efficient mechanism of association and disassociation based on signal level. 
	\item Appropriate values of bandwidth, loss, latency, and delays are applied to determine PER (packet error rate) and BER (bit error rate). 
\end{itemize}

The network simulation stack for CORNET 2.0 is based on Containernet, which extends Mininet-WiFi to allow Docker containers as mininet hosts and WiFi stations, and enables novel networking and cloud testbed features. Containernet extends the process-based virtualization of mininet to container-based virtualization.

\subsection{CORNET 2.0 Middleware}
The fundamental aspect of the proposed framework is to meaningfully interconnect the physical system simulator and network simulator coherently.  This is achieved through incorporating two primary functional blocks:
\begin{itemize}
    \item Establishing an end-to-end datapath for the nodes
    \item Mobility and time based synchronization between the two simulators
\end{itemize}

\begin{figure}[t]
\centering
	\includegraphics[width=0.9\columnwidth]{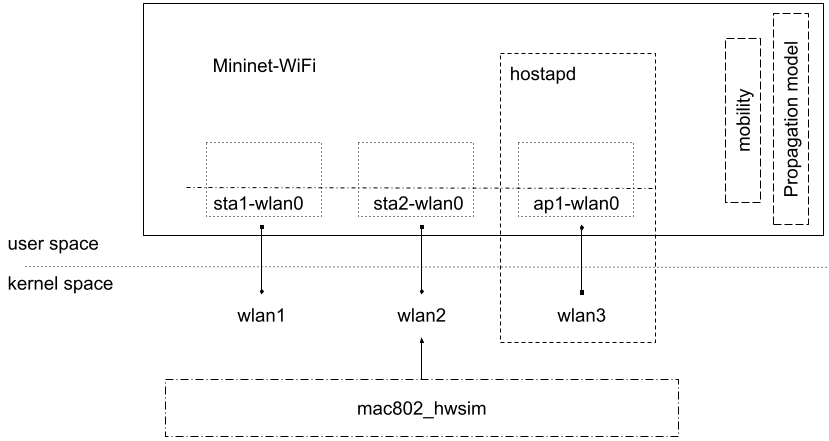}
	\caption{Mininet-WiFi Components\cite{mnwificomponents} }
	\label{fig:components}
\end{figure}

\subsubsection{End-to-end Datapath}
For every robot model in physics simulation a corresponding network host is created in mininet-wifi. We use a standard YAML configuration file to ensure one-to-one correspondence for a robot node. 
As shown in Fig. \ref{fig:components}, Mininet-Wifi uses kernel-space module \texttt{mac8021\_hwsim} to create virtual interfaces for mobile stations and access points. The virtual network is created by placing the robot host processes in different namespaces interconnected through virtual Ethernet (veth) pairs. CORNET utilizes this virtual network to establish an end-to-end data path between the robot nodes.
As shown in the data plane in Fig. \ref{fig:cornet2}, application developers can use any communication middleware, such as ROS, ROS2 for the interconnection of robots.

\subsubsection{Mobility and Time Synchronization}
The state information of each robot is obtained agnostic to the choice of the user application (ROS/ROS2). Gazebo uses boost ASIO to manage the communication layer and Google Protobufs for message passing and serialization library on named channels called topics via publishers. We use the native asyncio python bindings (pygazebo) for the Gazebo simulator. pygazebo supports publishing and subscribing to any Gazebo topic using a Python API.

In mininet Wifi, the constant position mobility model is enabled. An external program, i.e., the CORNET plugin, updates the position information from the state information obtained from the robot model in Gazebo. 

\begin{figure}[t]
	\centering
	\includegraphics[width=0.9\columnwidth]{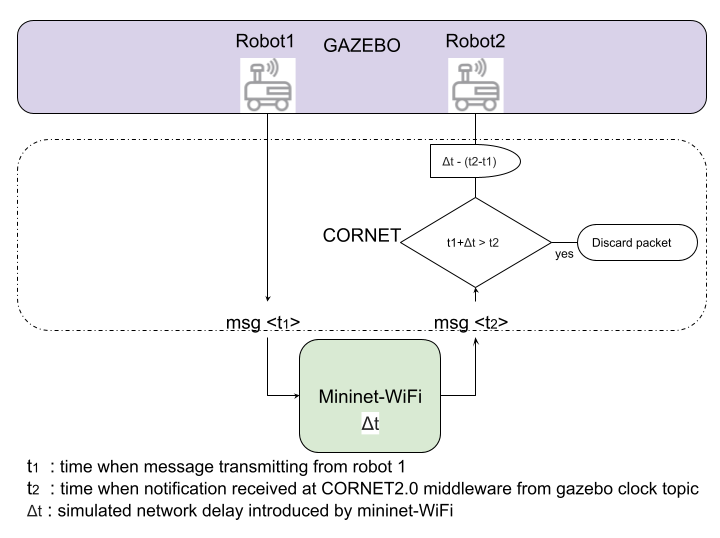}
	\caption{Time Synchronization}
	\label{fig:time_sync}
\end{figure}

Since both the simulators have different time models, it is critical for a co-simulation framework to ensure a common notion of time across the framework. In this framework, we use the same mechanism described in the CORNET 1.0\cite{acharya2020cornet} for time synchronization. We have used Gazebo's simulation time as the reference clock. Unlike in CORNET 1.0, the current implementation is not tightly coupled with ROS topics. As shown in Fig. \ref{fig:time_sync}, 
\begin{itemize}
    \item Every packet to and from the middleware (t1,t2) is time-stamped and released to the physical simulator only if the following condition is met. 
    
    \begin{itemize}
        \item If $(t2-t1) <\Delta t $,  the network event processing is faster than the physical system simulation. The CORNET middleware waits for $ (\Delta t - (t2-t1)) $ interval before releasing the packet.
        \item If $(t2-t1) > \Delta t $, the network event scheduler is slower than Gazebo time. In this case, we discard the packet as it missed the deadline.
    \end{itemize}
            
    \item In principle, the packet is delivered after \(\Delta Network - \Delta Physics \)  seconds, i.e., actual delay is measured network delay - time lapsed in the physics simulation for the same duration.   

\end{itemize}

\subsection{Discussion}
In summary, the CORNET 2.0 framework allows a broad spectrum of network-enabled robot simulation experiments.
\begin{itemize}
    \item It allows experimentation with network traffic shaping to smooth out the burst in traffic for better network behavior.
    \item It enables experiments in \textit{ad hoc} and mesh networks with better isolation for wireless interfaces.  
    \item It allows replaying of real network conditions based on actual communication traffic data logs. 
    \item It allows for configuring of arbitrary congestion profiles for the network traffic.
    \item It allows different channel models ranging from simple on/off links (disk models) to more complex statistical representations (fading models).
\end{itemize}

\section{Evaluation}
\begin{figure}[t]
	\centering
	
\subfloat[Without Synchronization]{
    \includegraphics[width=0.9\columnwidth]{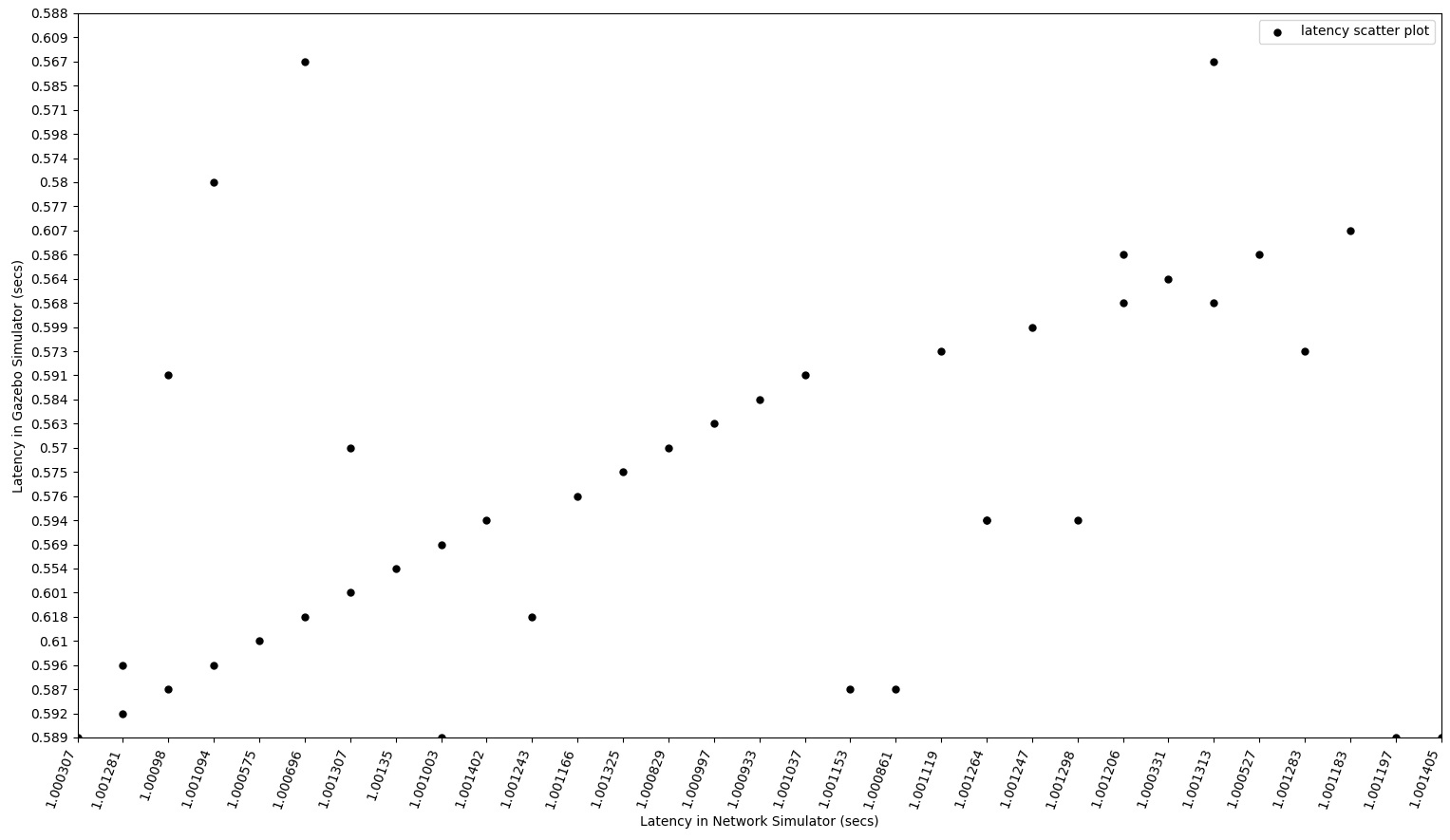} 
    \label{fig:without}
  }\\
  \subfloat[With Synchronization]{
   \includegraphics[width=0.9\columnwidth]{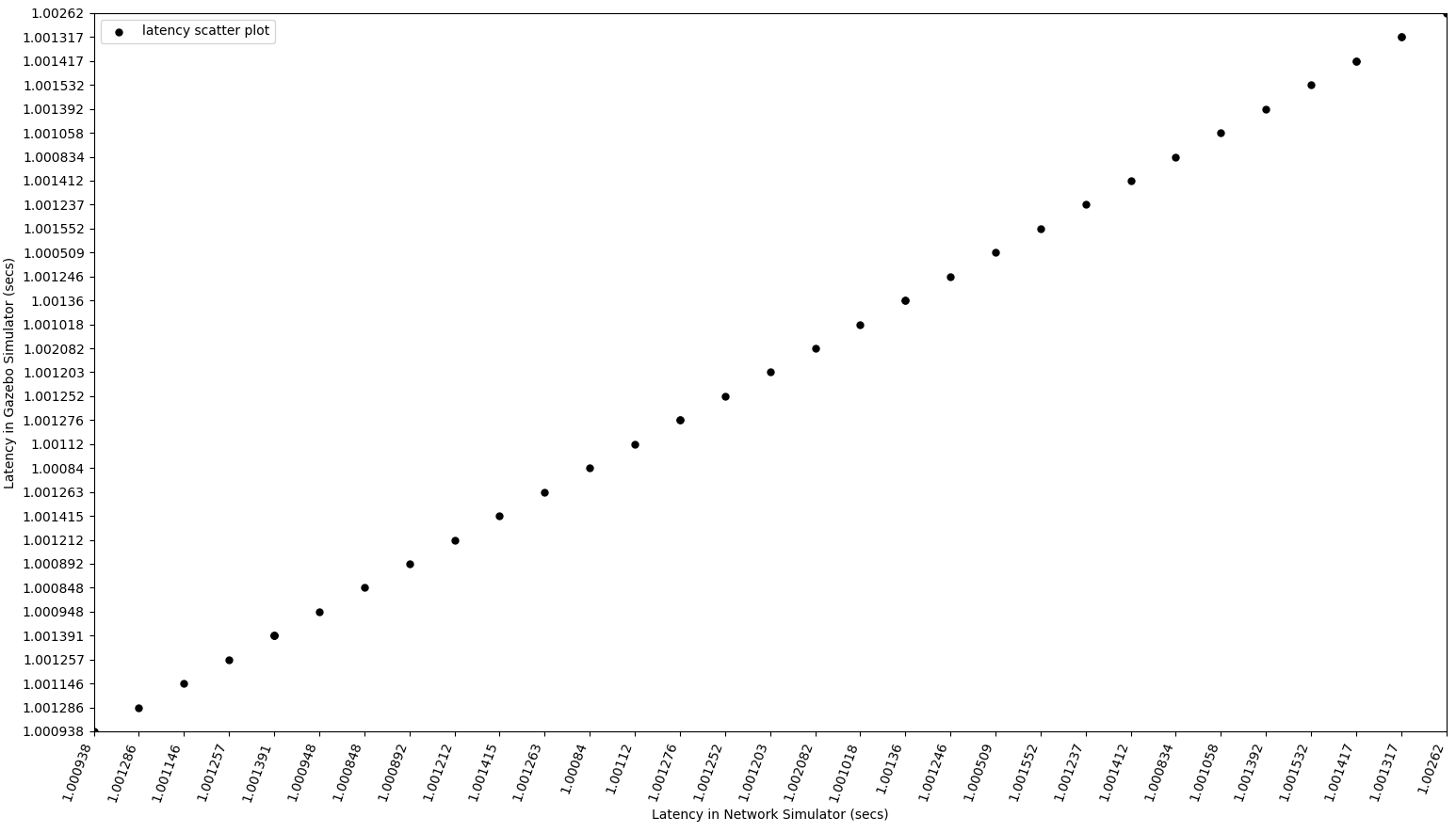}
    \label{fig:withSync}
  }
	\caption{Scatter Plot of latency in Network simulator vs. Gazebo simulator}
\end{figure}

In order to understand the benefits of time synchronization, we conduct two experiments, with and without time synchronization. In both, we simulate a network delay of one sec and measure the Gazebo time for the same duration. During these experiment, the latency of the Gazebo simulation was 0.49 - 0.65 of the real-world time. Fig. \ref{fig:without} and \ref{fig:withSync} shows the scatter plot of latency measured in gazebo vs network in the unsynchronized case and the synchronized case respectively. The results in Fig. \ref{fig:without}  does not indicates good correlation between the time of the packet measured within Gazebo and mininet. However, in Fig. \ref{fig:withSync}, with synchronization, the scatter plot shows a perfect alignment between the time measured as one moves from left to right. Synchronization plays a vital role, especially with the ever-changing real-time factor in the physical simulation, in ensuring the correctness of the simulation. This experiment indicates that CORNET is able to provide such string guarantees of time synchronization.

\begin{figure}[!h]
	\includegraphics[width=\linewidth]{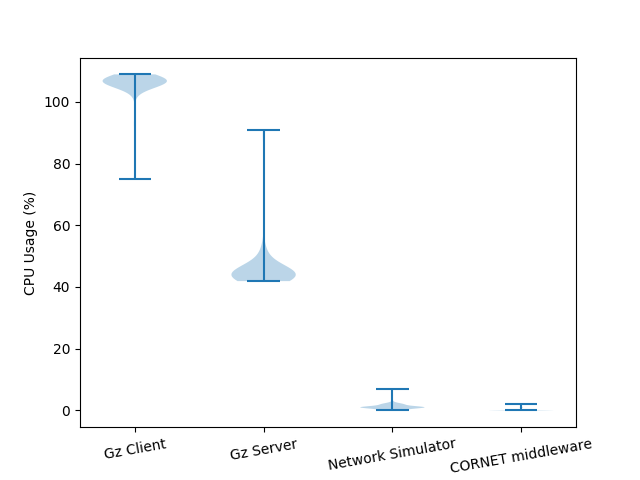}
	\caption{CPU Usage of individual processes in Simulation }
	\label{fig:process}
\end{figure}

\begin{figure}[!h]
	\includegraphics[width=\linewidth]{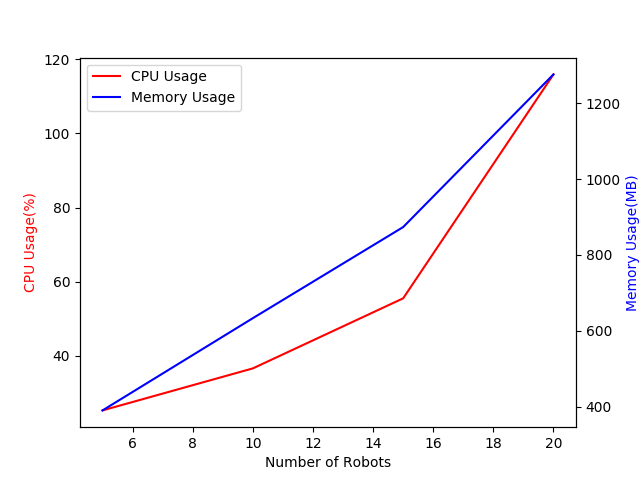}
	\caption{CPU and memory usage as a function of number of robots instantiated in the simulator}
	\label{fig:cpumem}
\end{figure}

We test the framework’s scalability by simulating multiple robots with the basic network configuration of a single access point and all the robots connected to a ROS master running in the access point.  The workstation used for the experiment has 32 GB of RAM and an Intel Core i7‑9700K processor. In Fig. \ref{fig:process}, we illustrate the average resource consumption of each component of the simulation framework. As expected gzclient, the Gazebo’s user interface that provides visualization and rendering of the simulation environment, consumes most of the CPU resources. In contrast, the CORNET middleware consumes a fraction of the CPU resources. For large-scale simulations, Gazebo allows the configuration to disable GUI and use only gzserver to realize the physics engine for the robots. We have disabled gzclient and instantiated up to 20 robots communicating using a WiFi network for the scaling experiment. Fig. \ref{fig:cpumem} shows both the CPU usage and memory usage as a function of the number of robots.  Our framework does not impose any restrictions on the number of robots, type of robot, and network topology. These results indicate that we can scale the simulation to 20 robots even on a modest computer.

CORNET2.0 inherits the limitation of CORNET where network simulator event processing has to be faster than the physical system simulator, i.e., Gazebo, which is the case in typical execution scenarios. CORNET2.0 achieves synchronization in one direction of simulation. Since in most of the user-machines real-time factor for the gazebo is less than 1. With more robots, the real-time factor falls drastically.

If the user wants a higher performance fidelity, in such cases, it is still possible to successfully emulate the network using a technique called VT-Mininet-Virtual-time\cite{yan2015vt}. The critical insight is to trade time with system resources by precisely scaling the time of interactions between network and physical devices by a factor of N, making the network appear to be N times faster from the viewpoints of the applications. 

The major upgrade in CORNET2.0 over CORNET is
\begin{itemize}
    \item It is not tightly coupled with the robotic application.
    \item It is independent of the choice of physical system simulator and network simulator.
    \item Robot communication middleware implementation is left to the user’s choice.
    \item It provides container-based virtualization for the robot nodes allowing for better scalability across compute clusters.
\end{itemize}

\subsection{CASE STUDY I: Mobile Robot }
\begin{figure}[!h]
	\includegraphics[width=\linewidth]{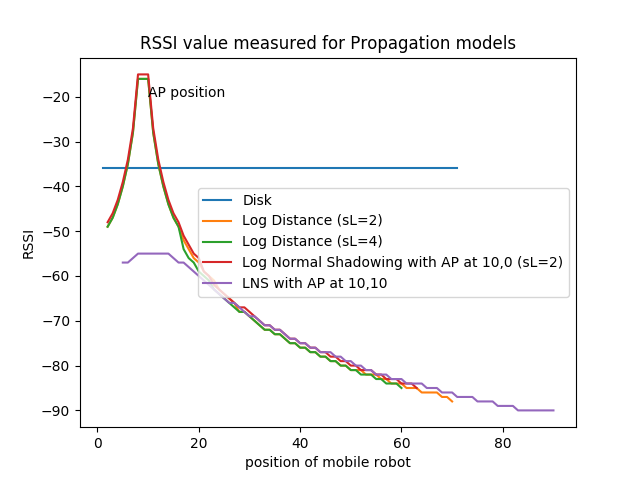}
	\caption{RSSI measured from a single access point }
	\label{fig:singleap}
\end{figure}

In this  scenario, We move a three-wheeled robot in an empty world,  
\begin{itemize}
    \item [a.] connected to single access point
    \item [b.] connected to multiple access points(Handover)
\end{itemize}
Fig. \ref{fig:singleap} shows how the RSSI value measured in network simulator changes with respect to the position of the mobile robot node from the physics simulator for various propagation models. In all cases, the robot in the gazebo simulation moves from (0,0) to (100,0) in a straight line. Except in the last case, as shown in the Fig. \ref{fig:singleap}, the access point is located at (10,0). 

When the robot node moves closer to the access point, the signal quality improves, and when it moves away, the signal quality degrades. 

These trends in the graph clearly show that with changes in the propagation model parameters (SL), the RSSI value measurement of the mobile robot is affected, i.e., more the system loss value lesser the range of signal connection.  Changes in the position of the access point also reflect the RSSI measurements (purple trace). 
\begin{figure}[!h]
	\includegraphics[width=\linewidth]{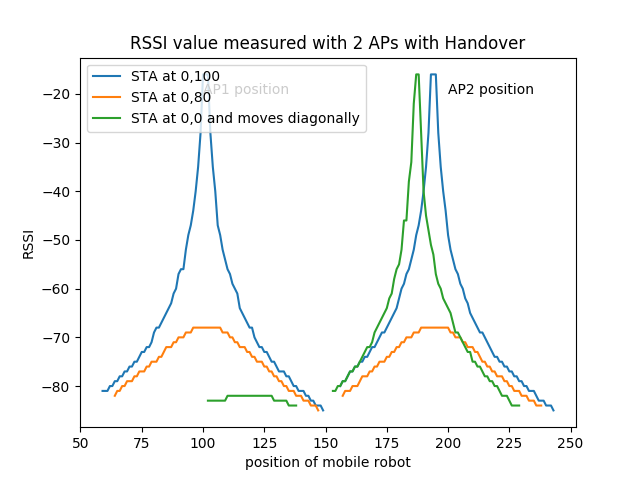}
	\caption{RSSI measured with two access points (handover case) }
	\label{fig:2ap}
\end{figure}
In the second case, we repeat the experiment with two access points and observe the effects of the handovers. In this experiment, the first access point is at (100,100), and the second access point is at (200,100), and in the first iteration(blue), the robot node moves from (0,100) to (300,100) in a straight line. In second iteration(orange),the robot node moves from (0,80) to (300,80) in a straight line. Furthermore, in the third iteration (green), the robot node moves from (0,0) to (300,100). 
The data in Fig. \ref{fig:2ap} clearly illustrates a similar trend as the first case. Interestingly, when the robot is in the range of both access points i.e., at (150,0), the handover mechanism occurs. The robot is not associated with any access point during this time, resulting in a measured RSSI value of zero.   
In this experiment, the main objective is to highlight the relationship between network nodes and physics node co-simulated. 

\subsection{CASE STUDY II: Controllability of inverted pendulum}
\begin{figure}[!h]
	\includegraphics[width=\linewidth]{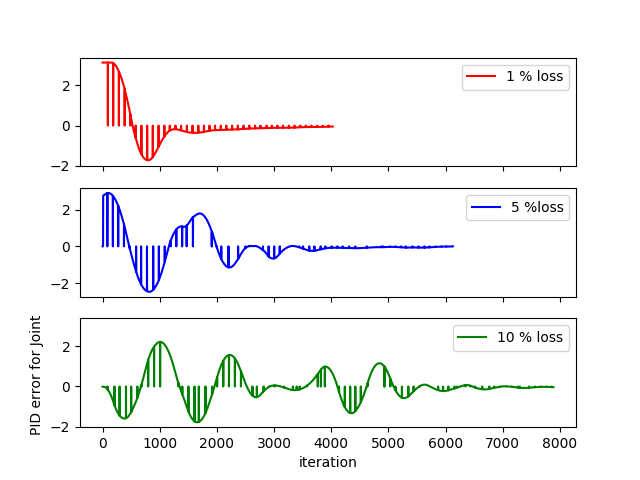}
	\caption{Networked PID Controller error vs. Packet loss }
	\label{fig:ipc}
\end{figure}

In this scenario, we control an inverted pendulum with a PID controller from a remote mobile station. The topology is such that the inverted pendulum robot model is connected to the access point, and we try to control it over a lossy network.  Fig. \ref{fig:ipc} illustrates controller position error versus packet loss in a simulated channel. We see that the time to converge increases with packet loss.   
This use case demonstrates the stochastic nature of the medium-induced artifacts like packet loss, jitter, and time-varying delays on developing algorithms for networked control systems.  

Note: For case studies I and II, the control algorithm is developed using ROS, and the source code is available at github \cite{ipc}.  

\subsection{CASE STUDY III: Teleoperation of Warehouse Robots}
\begin{figure}[!h]
	\includegraphics[width=\linewidth]{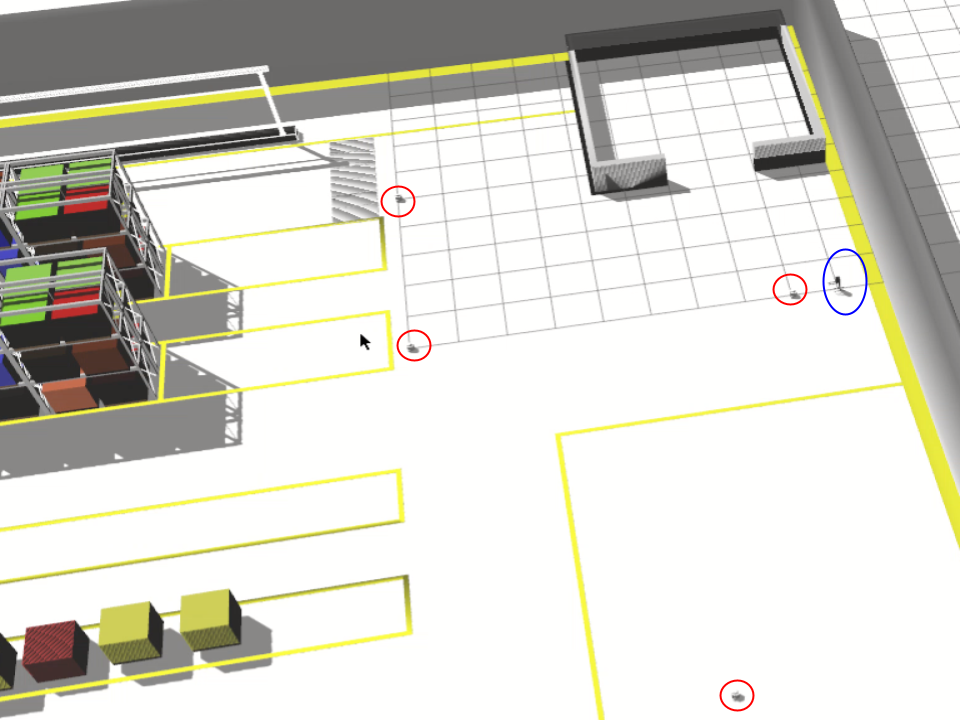}
	\caption{Teleoperation of Warehouse robots }
	\label{fig:tele}
\end{figure}

In this scenario, we created a warehouse environment and demonstrated the use-case of the teleoperation of robots from a central control station, as illustrated in Fig. \ref{fig:tele}. We have used ROS2 middleware to showcase the end-to-end integration of CORNET 2.0 with gazebo and Containernet. As shown in Fig. \ref{fig:tele}, Gazebo instantiates robots and the corresponding world environment for the warehouse use case (Red circles marks the robot positions and blue circle for access point). 
The teleoperation node connects to the access point via an ethernet link. It uses ROS2 package teleop\_twist\_keyboard for keyboard-based teleoperation of the robot nodes connected over WiFi. As long as the robots are in the access point’s range, the teleoperation can control the robots and send commands over the simulated network.   

This use case demonstrates the end-to-end interaction between the two simulation environments for a multi-robot scenario.
  
The video demonstration of this use case is provided here(https://www.youtube.com/watch?v=cJScgPS5Pz0). 

\section{Conclusion}
In this work, we have presented CORNET2.0, a framework for simulating Multi-Robot Systems that combines physics and network simulators. We have evaluated this framework’s usability for two different application middleware implementations, mainly ROS and ROS2. We have maintained the key features of CORNET, i.e., time and mobility synchronization, and generalized the implementation to extend the applicability to any robotic framework. We have demonstrated the scalability of framework to handle an increasing number of robots. As future work, we are planning to support the hardware in the loop simulation experiments.

\section*{Acknowledgment}
We would like to thank ARTPARK, RBCCPS, and Nokia Research Fellowship for their constant support. Furthermore, we are very grateful to the networked warehouse robotics project members especially Naveen, Shubham and Parv for testing the framework rigorously and helping to improve the framework.

\bibliographystyle{abbrv}
\bibliography{simple}

\end{document}